\documentclass[twoside,twocolumn,11pt]{article}
\def\anon{0}
\def\lineno{0}
\usepackage{fancyhdr}
\usepackage{color}
\usepackage[color=black,opacity=1,angle=0,scale=1]{background}
\usepackage[pscoord]{eso-pic}

\newcommand{\placetextbox}[3]{
  \setbox0=\hbox{#3}
  \AddToShipoutPictureFG*{
    \put(\LenToUnit{#1\paperwidth},\LenToUnit{#2\paperheight}){\vtop{{\null}\makebox[0pt][c]{#3}}}%
  }%
}%

\backgroundsetup{
  contents={%
    \placetextbox{0.5}{0.1}{To appear in www.WSCG.eu proceedings/journal 2017}    
    \placetextbox{0.5}{0.085}{p. \thepage}%
  }
}

\makeatletter
\newsavebox{\@linebox}
 \savebox{\@linebox}[3em][t]{\parbox[t]{3em}{%
   \@tempcnta\@ne\relax
   \loop{\underline{
     \the\@tempcnta}}\\
     \advance\@tempcnta by \@ne\ifnum\@tempcnta<51\repeat}}

\fancypagestyle{numberstyle}{
 \fancyhead{}
 \fancyfoot{}
\if\lineno1
 \fancyhead[LE]{
     \begin{picture}(0,0)%
      \put(-20,-15){\usebox{\@linebox}}%
      \put(480,-15){\usebox{\@linebox}}%
      \end{picture}}

 \fancyhead[LO]{
    \begin{picture}(0,0)%
      \put(-30,-15){\usebox{\@linebox}}%
      \put(470,-15){\usebox{\@linebox}}%
     \end{picture}}
\fancyfoot[C]{\scriptsize \thepage}
\else
\fi
\makeatother
}
\usepackage{marginnote}
\usepackage{graphicx}
\usepackage{subfig}
\usepackage{wscg}           
\usepackage[hidelinks]{hyperref}
\RequirePackage{ifpdf}
\RequirePackage{amsmath}
\RequirePackage{amsfonts}
\RequirePackage{amssymb}
\RequirePackage{amsxtra}

\title{Interpatient Respiratory Motion Model Transfer for Virtual Reality Simulations of Liver Punctures}

\if\anon1
\author{
\parbox{0.25\textwidth}{\centering
First Author\\[1mm]
author's affiliation\\
1st line of address\\
2nd line of address\\
Country (ZIP) code, City, State\\[1mm]
e@\\mail
}
\hspace{0.05\textwidth}
\parbox{0.25\textwidth}{\centering
Second Author\\[1mm]
author's affiliation\\
1st line of address\\
2nd line of address\\
Country (ZIP) code, City, State\\[1mm]
e@\\mail
}
\hspace{0.05\textwidth}
\parbox{0.25\textwidth}{\centering
Third Author\\[1mm]
author's affiliation\\
1st line of address\\
2nd line of address\\
Country (ZIP) code, City, State\\[1mm]
e@\\mail
}
}
\else 
\author{
\parbox{0.25\textwidth}{\centering
Andre Mastmeyer\\[1mm]
Univ. of Luebeck\\
Inst. of Med. Inform.\\
Ratzeburger Allee 160\\
23568 Luebeck, Germany\\[1mm]
mastmeyer@imi.uni-luebeck.de
}
\hspace{0.05\textwidth}
\parbox{0.25\textwidth}{\centering
Matthias Wilms\\[1mm]
Univ. of Luebeck\\
Inst. of Med. Inform.\\
Ratzeburger Allee 160\\
23568 Luebeck, Germany\\[1mm]
wilms@imi.uni-luebeck.de
}
\hspace{0.05\textwidth}
\parbox{0.25\textwidth}{\centering
Heinz Handels\\[1mm]
Univ. of Luebeck\\
Inst. of Med. Inform.\\
Ratzeburger Allee 160\\
23568 Luebeck, Germany\\[1mm]
handels@imi.uni-luebeck.de
}
}
\fi 

\usepackage{url}
\makeatletter
\def\Uslash{\mathbin{\mathchar`\/}\@ifnextchar{/}{\kern-.15em}{}}
\g@addto@macro\UrlSpecials{\do \/ {\Uslash}}
\def\Ucolon{\mathbin{\mathchar`:}\@ifnextchar{/}{\kern-.1em}{}}
\g@addto@macro\UrlSpecials{\do : {\Ucolon}}
\makeatother

\begin{document}

\twocolumn[{\csname @twocolumnfalse\endcsname

\maketitle  
\thispagestyle{numberstyle}

\begin{abstract}\noindent
Current virtual reality (VR) training simulators of liver punctures often rely on static 3D patient data and use an unrealistic (sinusoidal) periodic animation of the respiratory movement.
Existing methods for the animation of breathing motion support simple mathematical or patient-specific, estimated breathing models. However with personalized breathing models for each new patient, a heavily dose relevant or expensive 4D data acquisition is mandatory for keyframe-based motion modeling.
Given the reference 4D data, first a model building stage using linear regression motion field modeling takes place. Then the methodology shown here allows the transfer of existing reference respiratory motion models of a 4D reference patient to a new static 3D patient. This goal is achieved by using non-linear inter-patient registration to warp one personalized 4D motion field model to new 3D patient data.
This cost- and dose-saving new method is shown here visually in a qualitative proof-of-concept study. 
\end{abstract}

\subsection*{Keywords}
Virtual Reality, Liver Puncture Training, 4D Motion Models, Inter-patient Registration of Motion Models

\vspace*{1.0\baselineskip}
}]

\section{INTRODUCTION}

\copyrightspace

The virtual training and planning of minimally invasive surgical interventions with virtual reality simulators provides an intuitive, visuo-haptic user interface for the risk-sensitive learning and planning of interventions. 
The simulation of liver punctures has been an active research area for years \cite{G53-01,G53-01a,G53-04}.

\begin{figure*}
  \centering		    
	\includegraphics[bb=0bp 0bp 1622bp 586bp,clip,height=5.8cm]{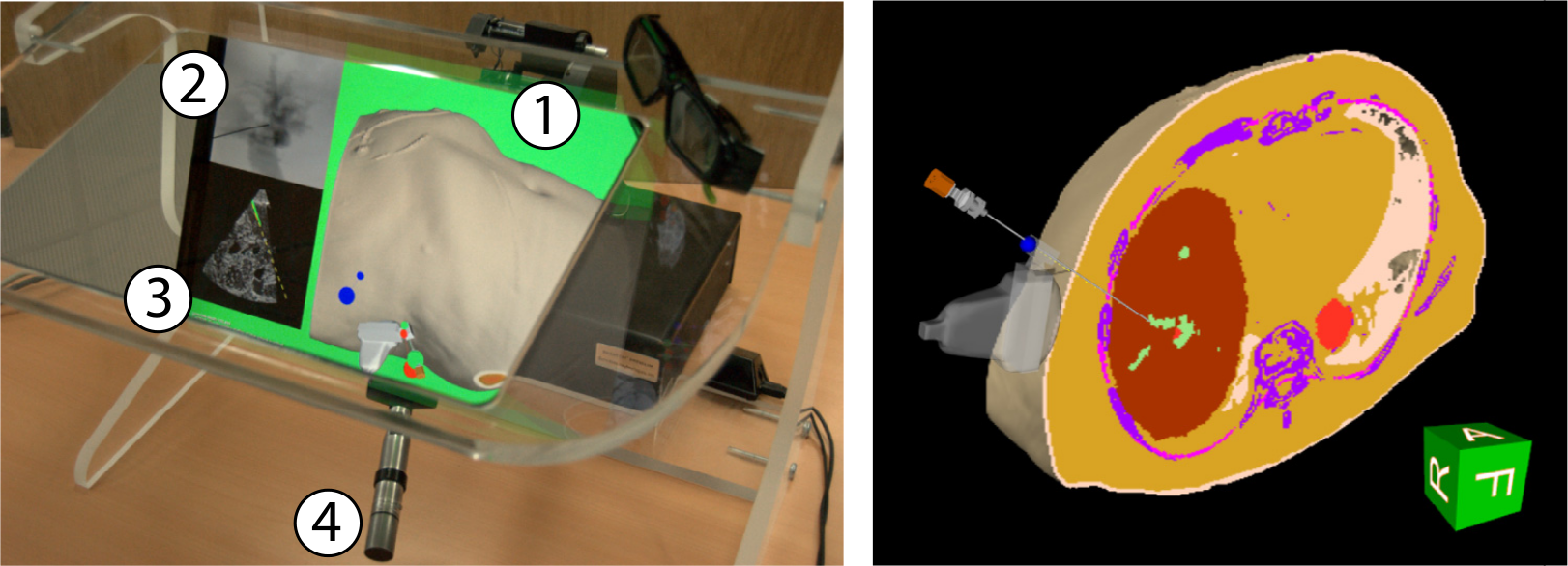}
\caption{\label{G53-fig1}Left: Hardware: (1) Main stereo rendering window with successful needle insertion into a target, (2) fluoroscopic simulation, (3) Ultrasound simulation, (4) haptic device handle. Right: Main rendering window displaying oblique cut and color-coded patient structures just before needle insertion into a targeted bile duct (green).}
\end{figure*}

{\color{black}Obviously first, the stereoscopic visualization of the anatomy of the virtual patient body is important \cite{G53-06}. Second, the haptic simulation of the opposing forces through the manual interaction, rendered by haptic input and output devices, with the patient is key \cite{G53-05}. Third in recent developments, the simulation of the appearance and forces of the patient's breathing motions is vital \cite{G53-20,G53-04}.}

The previously known VR training simulators usually use time invariant 3D patient models. A puncture of the spinal canal can be simulated sufficiently plausibly by such models. In the thoracic and upper abdominal region, however, respiratory and cardiac movements are constantly present. In the diaphragm area at the bottom of the lungs just above the liver, breathing movement differences in the {\color{black}longitudinal z} direction of up to 5 cm were measured \cite{G53-07}. Now for 4D animation, the necessary data consists of a single 3D CT data set and a mathematical or personalized animation model.
Our 
aim here is to incorporate these physiological-functional movements into realistic modeling in order to offer the user a more realistic visuo-haptic VR puncture simula\-tion. This means also to take into account the intra- and intercycle variability (hysteresis, variable amplitude during inhalation / exhalation).

{\color{black}A major interest and long term goal of virtual and augmented reality is the planning \cite{G53-19} and intra-operative navigation assistance \cite{G53-18}. However, in these works breathing motion is not incorporated or applicability limits by neglecting breathing motion in terms of minimal tumor size are given \cite{G53-18}.}
Published approaches from other groups \cite{G53-02,G53-03} model only a sinusoidal respiratory motion without hysteresis and {\color{black}amplitude variation. First steps in the direction of a motion model building framework were taken by our group \cite{G53-16}.} Accurate simulation of respiratory motion depending on surrogate signals is relevant e.g. in fractionated radiotherapy.
However, since a patient-specific 4D volume data set is required for personalized breathing model building and its acquisition is associated with a high radiation dose with 4D-CT ($\geq$ 20-30 mSv (eff.)), our approach is
the transfer of existing 4D breathing models to new 3D patient data.
For comparison, the average natural background radiation is approximately 2.1 mSv (eff.) per year\footnote{\href{http://www.bfs.de/EN/topics/ion/environment/natural-radiation-exposure/natural-radiation-exposure_node.html}{Intercontinental flight max. 0.11 mSv (eff.)}}.

On the other hand, there is no medical indication to acquire 4D CT data just for training purposes and model building from 4D MR data to be included is unjustifiable for cost reasons.

In this paper, we present a feasibility study with first qualitative results for the transfer of an existing 4D breathing model 
\cite {G53-11} to static 3D patient data, in which only a 3D CT covering chest and upper abdomen at maximum inhalation is necessary (approximately 2- 13 mSv (eff.))\footnote{\href{https://static.healthcare.siemens.com/siemens_hwem-hwem_ssxa_websites-context-root/wcm/idc/groups/public/@global/@imaging/@ct/documents/download/mdaw/mtm1/~edisp/ct_somatom_definition_as_brochure-00032845.pdf}{Siemens Somatom Definition AS}}.

\section{RECENT SOLUTION}\label{sec:recent}\label{G53-sec:vrsimulator}
The existing solution requires a full 4D data set acquisition for each new patient.
In \cite {G53-01,G53-09,G53-10,G53-12}, concepts for a 3D VR simulator and efficient patient modeling for the training of different punctures (e.g. liver punctures) have already been presented, see Fig.~\ref{G53-fig1}. A Geomagic Phantom Premium 1.5 HighForce is used for the manual control and haptic force feedback of virtual surgical instruments.  Nvidia shutter glasses and a stereoscopic display provide the plausible rendering of the simulation scene. This system uses time invariant 3D CT data sets as a basis for the patient model.
In case of manual interaction with the model, tissue deformation due to acting forces of the instruments are represented by a direct visuo-haptic volume rendering method.

New developments of VR simulators
\cite {G53-01a} allow a time-variant 4D-CT data set to be used in real time for the visualized patient instead of a static 3D CT data set. The respiratory movement can be visualized visuo-haptically as a keyframe model using interpolation or with a flexible linear regression based breathing model as described below.

\section{PROPOSED SOLUTION}
The new solution requires only a 3D data set acquisition for each new patient. 
\subsection{Modeling of Breathing Motion}
\label{G53-sec:model}
Realistic, patient-specific modeling of breathing motion in \cite {G53-01a}  relies on a 4D CT data set covering one breathing cycle. It consists of $\mathrm{N}_{phases}$ phase 3D images indexed by $j$. Furthermore, a surrogate signal (for example, acquired by spirometry) to parametrize patient breathing in a low-dimensional manner is necessary.

We use a measured spirometry signal $ v (t) $ [ml] and its temporal derivative in a composite surrogate signal: $
(v (t), v '(t))^T.
$
This allows to describe different depths of breathing and the distinction between inhalation and exhalation (respiratory hysteresis). We assume linearity between signal and motion extracted from the 4D data. {\color{black}First, we use the 'sliding motion'-preserving approach from \cite{G53-14} for $\mathrm{N}_{phases}-1$ intra-patient inter-phase image registrations to a selected reference phase $j_{ref}$:
\begin{align}\label{eq:intra} \nonumber
&\varphi^{pat4D}_{j}=\\
&\underset{\varphi}{\mathrm{argmin}}\left(D_{NSSD}\left[I^{pat4D}_{j}, I^{pat4D}_{j_{ref}} \circ \varphi\right]
+\alpha_S \cdot R_{S}(\varphi)\right),\\ \nonumber
&\qquad \qquad  j\in \{1,..,j_{ref}-1,j_{ref}+1,..,\mathrm{N}_{phases}\},
\end{align}
where a distance measure $ D_{NSSD} $ (normalized sum of voxel-wise squared differences \cite{G53-17}) and a specialized regularization $ R_{S} $} establishes smooth voxel correspondences except in the pleural cavity where discontinuity is a wished feature \cite{G53-14}.
Based on the results, the coefficients $ {a ^ {pat4D}_{1..3}} $ are estimated as vector fields over the positions $ \mathbf {x} $. The personalized breathing model then can be stated as a linear multivariate regression \cite{G53-11}:
\begin{align}\label{eq:model4D} \nonumber
\hat{\varphi}^{pat4D}(\mathbf{x},t)=
&a^{pat4D}_1(\mathbf{x})\cdot v(t)~+\\ \nonumber
&a^{pat4D}_2(\mathbf{x})\cdot v'(t)~+\\
&a^{pat4D}_3(\mathbf{x}),~~\mathbf{x} \in \Omega_{pat4D}.
\end{align}
Thus, a patient's breathing state can be represented by a previously unseen breathing signal: Any point in time $ t $ corresponds to a shifted reference image $I^{pat4D}_{j_{\mathrm{ref}}} \circ \hat{\varphi}^{pat4D}(\mathbf{x},t)$.
Equipped with a real-time capable rendering technique {\color{black}via ray-casting with bent rays (see \cite {G53-01a} for full technical details), the now time variant model-based animatable CT data $I^{pat4D}_{j_{\mathrm{ref}}}$ can be displayed in a new variant of the simulator and used for training. The rays are bent by the breathing motion model and this conveys the impression of an animated patient body, while being very time efficient (by space-leaping and early ray-termination) compared to deforming the full 3D data set for each time point and linear ray-casting afterwards \cite{G53-01a}.} 

\subsection{Transfer of Existing Respiratory Models to new, static Patient Data}
\label{G53-sec:transfer}
Using the method described so far, personalized breathing models can be created, whose flexibility is sufficient to approximate the patients' breathing states, which are not seen in the observation phase of the model formation.

However, the dose-relevant or expensive acquisition of at least one 4D data set has thus far been necessary for each patient.

Therefore, here we pursue the idea to transfer a readily built 4D reference patient breathing model to new static patient data $ pat3D $ and to animate it in the VR simulator described in Sec.~\ref {G53-sec:vrsimulator}.

For this purpose, it is necessary to correct for the anatomical differences between the reference patient with the image data $I^{pat4D}_{j_{\mathrm{ref}}}$ and the new patient image data $I^{pat3D}_{\mathrm{ref}}$ based on a similar breathing phase. This is achieved, for example, by a hold-breath scan ($ \mathrm{ref} $) in the maximum inhalation state, which corresponds to a certain phase $ j_{\mathrm{ref}} $ in a standardized 4D acquisition protocol. A nonlinear inter-patient registration $ \varphi (\mathbf {x}): \Omega_{pat3D} \rightarrow \Omega_{pat4D} $ with minimization of a relevant image distance $ D $ ensures the necessary compensation \cite{G53-09,G53-10}:
\begin{align}\label{eq:inter} \nonumber
&\varphi^{pat3D \rightarrow pat4D}_{j_{\mathrm{ref}}}=\\ 
&\underset{\varphi}{\mathrm{argmin}}\left(D_{SSD}\left[I^{pat3D}_{\mathrm{ref}}, I^{pat4D}_{j_{\mathrm{ref}}} \circ \varphi\right]
+\alpha_{D} \cdot R_{D}(\varphi)\right),
\end{align} 
where a distance measure $ D_{SSD} $ (sum of squared voxel-wise differences) and a diffusive non-linear regularization $ R_{D} $ establishes smooth inter-patient voxel correspondences.
On both sides, the breathing phase 3D image of maximum inhalation is selected as the reference phase (ref).
The distance measurement can be selected according to the modality and quality of the image data.
The transformation $\varphi^{pat3D \rightarrow pat4D}_{j_{\mathrm {ref}}}$, which is determined in the nonlinear inter-patient registration, can now be used to warp the intra-patient inter-phase deformations of the reference patient $\varphi^{pat4D}_{j} $ as a plausible estimate $ {\varphi} ^ {pat3D}_{j} $ ($ j \in \{1, \ldots, n \} $; $ \circ $: right to left):
\begin{align}\nonumber
&{\varphi}^{pat3D}_{j} =\\ 
&\left(
\varphi^{pat3D \rightarrow pat4D}_{j_{\mathrm{ref}}} 
\right)^{-1}
\circ 
\varphi^{pat4D}_{j} \circ 
\varphi^{pat3D \rightarrow pat4D}_{j_{\mathrm{ref}}}.
\end{align}
The approach for estimating the respiratory motion for the new patient can now be applied analogously to the reference patient (see Sec.~\ref {G53-sec:model}). With a efficient regression method \cite{G53-11}, the breathing movement of virtual patient models, which are only based on a comparatively low dose of acquired 3D-CT data, can be plausibly approximated:
\begin{align}\label{eq:model3D} \nonumber
\hat{\varphi}^{pat3D}(\mathbf{x},t)=\nonumber
&{a}^{pat3D}_1(\mathbf{x})\cdot v(t)~+\\ \nonumber
&{a}^{pat3D}_2(\mathbf{x})\cdot v'(t)~+\\
&{a}^{pat3D}_3(\mathbf{x}),~~\mathbf{x} \in \Omega_{pat3D}.
\end{align}
Optionally, simulated surrogate signals $ v (t) $ 
can be used for the 4D animation of 3D CT data. Simple alternatives are to use the surrogate signal of the reference patient or also a (scaled) signal of the new patient $pat3D$, which can simply be recorded with a spirometric measuring device without new image acquisition.

\begin{figure*}
  \centering		    
	\subfloat[Field of view difference between reference patient $ pat4D $ (turquoise) and target patient $ pat3D $ (yellow).\label{sfig:pat3D4D}]{\includegraphics[bb=0bp 0bp 482bp 539bp,clip,height=7.5cm]{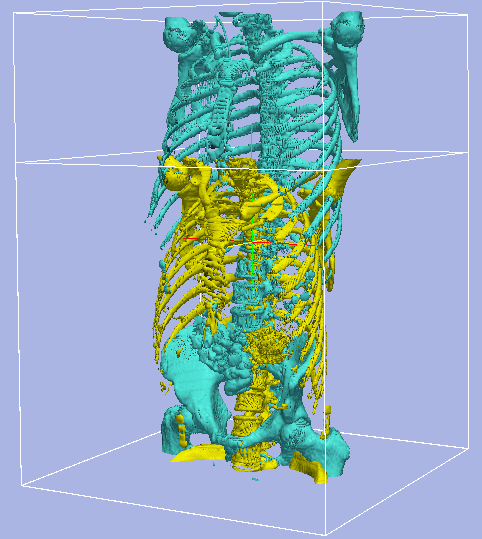}}
    \subfloat[Selected times of the spirometry signal from $pat3D $ (blue).\label{sfig:spiro}]{\includegraphics[height=7.5cm]{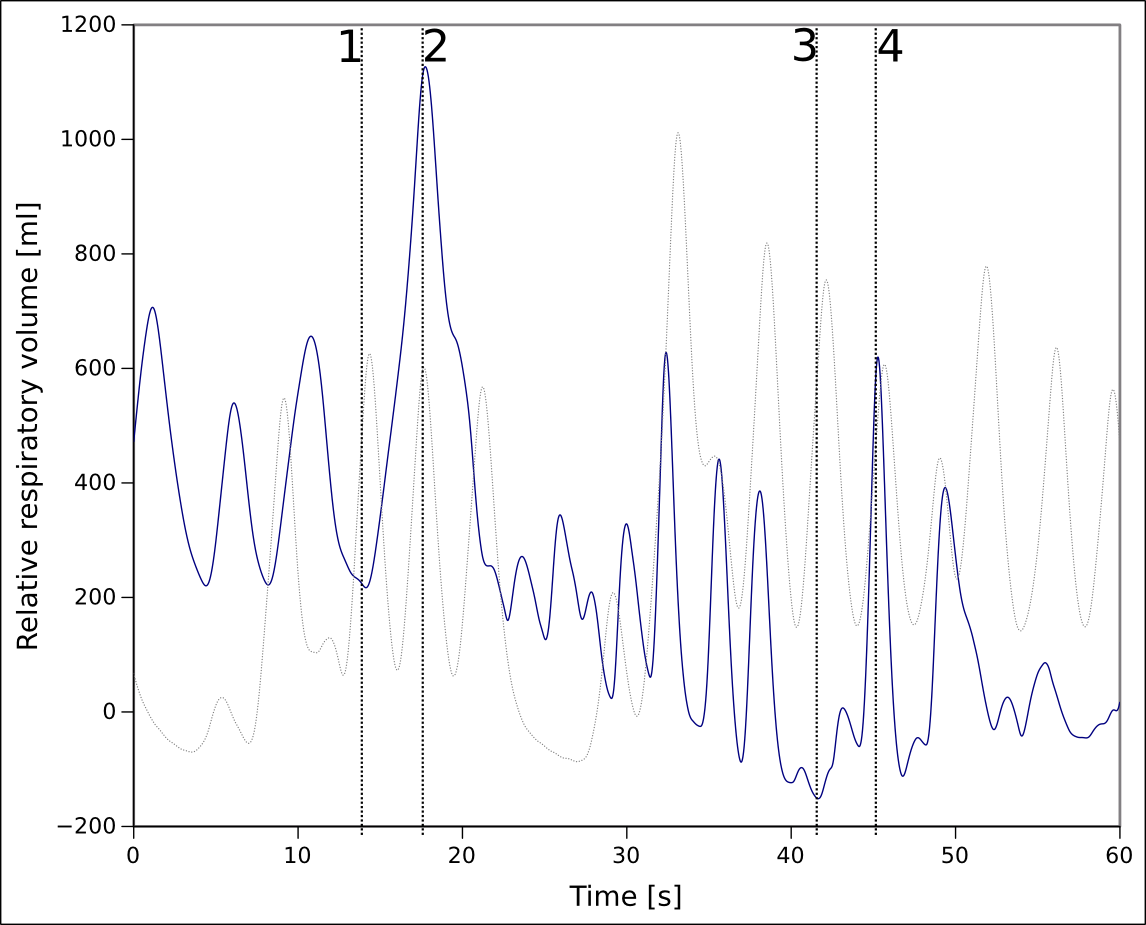}}
	\caption{(a) Field of views and (b) respiratory signals of the patients $pat4D$ (gray dashed) vs. $pat3D$.
}
\label{G53-img:FoV-motionsig}
\end{figure*}
\begin{figure*}
  \centering	
    \subfloat[First time point.\label{sfig:}]{\includegraphics[bb=0bp 0bp 317bp 287bp,clip,height=7.2cm]{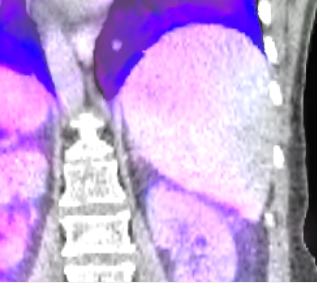}}~
    \subfloat[Second time point.\label{sfig:}]{\includegraphics[bb=0bp 0bp 317bp 283bp,clip,height=7.2cm]{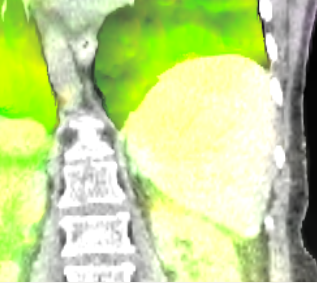}}\\
    \subfloat[Third time point.\label{sfig:}]{\includegraphics[bb=0bp 0bp 317bp 283bp,clip,height=7.2cm]{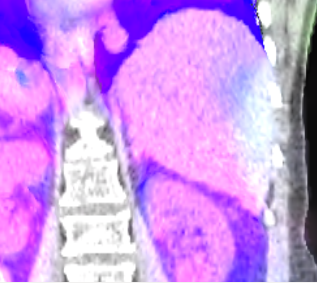}}~
    \subfloat[Fourth time point.\label{sfig:}]{\includegraphics[bb=0bp 0bp 317bp 283bp,clip,height=7.2cm]{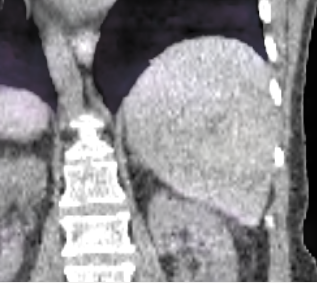}}
\\\subfloat{\includegraphics[bb=0bp 0bp 136bp 159bp,height=2cm]{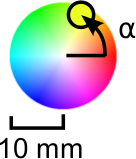}}
	\caption{Field of view, respiratory signal and coronal views with overlayed motion field to the CT data of the patients $ pat4D $ (a-d). The color wheel legend below indicates the direction of the motion field.
}
\label{G53-img:motionct4D}
\end{figure*} 

\begin{figure*}
  \centering	
    \subfloat[First time point.\label{sfig:}]{\includegraphics[bb=0bp 0bp 317bp 283bp,height=7.2cm]{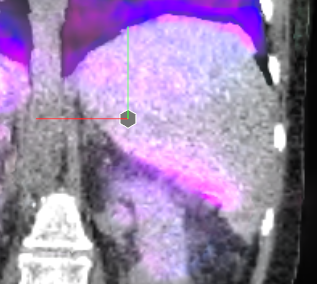}}~
    \subfloat[Second time point.\label{sfig:}]{\includegraphics[bb=0bp 0bp 317bp 284bp,height=7.2cm]{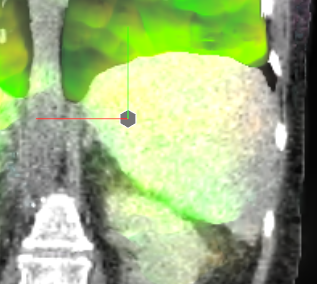}}\\
    \subfloat[Third time point.\label{sfig:diaArt}]{\includegraphics[bb=0bp 0bp 317bp 284bp,height=7.2cm]{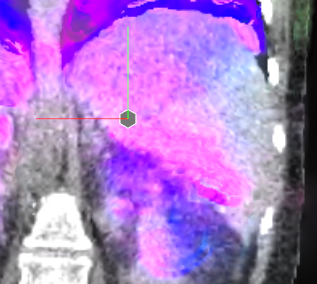}}~
    \subfloat[Fourth time point.\label{sfig:}]{\includegraphics[bb=0bp 0bp 317bp 284bp,height=7.2cm]{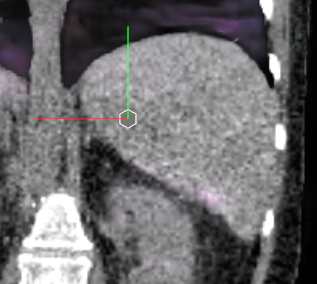}}
    \\\subfloat{\includegraphics[bb=0bp 0bp 136bp 159bp,height=2cm]{G53-fig10.png}}
	\caption{Coronal views with overlayed motion field to the CT data of the patient $ pat3D $ (a-d) deformed with the model of $ pat4D $. The color wheel legend below indicates the direction of the motion field.
}
\label{G53-img:motionct3D}
\end{figure*}

\begin{figure*}
    \subfloat[First time point.\label{sfig:}]{\includegraphics[bb=0bp 0bp 316bp 255bp,height=7.2cm,trim={1.1cm 0 0 0},clip]{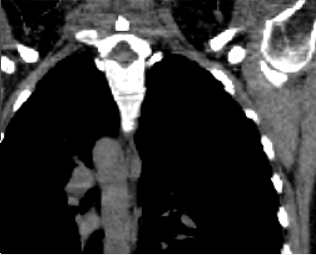}}~
    \subfloat[Second time point.\label{sfig:}]{\includegraphics[bb=0bp 0bp 316bp 255bp,height=7.2cm,trim={1.1cm 0 0 0},clip]{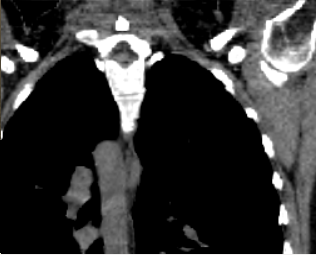}}\\
    \subfloat[Third time point.\label{sfig:ribArt}]{\includegraphics[bb=0bp 0bp 316bp 255bp,height=7.2cm,trim={1.1cm 0 0 0},clip]{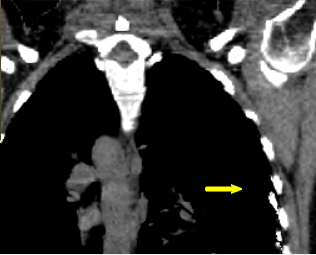}}~
    \subfloat[Fourth time point.\label{sfig:}]{\includegraphics[bb=0bp 0bp 316bp 255bp,height=7.2cm,trim={1.1cm 0 0 0},clip]{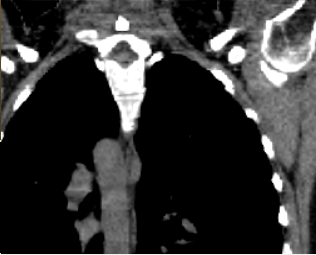}}
	\caption{Upper thorax coronal views of the animated CT data of the patient $ pat3D $ (a-d) deformed with the model of $ pat4D $. Rib artifacts are indicated by the yellow arrow in (c).
}
\label{G53-img:motionArt}
\end{figure*}

\section{EXPERIMENTS AND RESULTS}

We performed a qualitative feasibility study, results are animated in the 4D VR training simulator \cite {G53-01a}.

For the 4D reference patient, a 4D-CT data set of the thorax and upper abdomen with 14 respiratory phases (512$^2 \times $ 462 voxel to 1$^3$ mm) and a spirometry signal $ v (t) $ were used (Fig.~\ref{G53-img:FoV-motionsig}).
The new patient is represented only by a static 3D CT data set (512$^2 \times$ 318 voxel to 1$^3$ mm).

All volume image data was reduced to a size of 256$^3$ voxel due to the limited graphics memory of the GPU used (Nvidia GTX 680 with 3 GB RAM).

According to Eq.~\ref{eq:intra} we first perform the intra-patient inter-phase registrations to a chosen reference phase $j_{ref}$.

The registrations from Eqs. \ref{eq:intra} and \ref{eq:inter} use weights $ \alpha_{S} = 0.1 $ and $\alpha_D=1$ for the regularizers $R_{S}$ and $R_{D}$. In both registration processes, the phase with maximum inhalation is used as the reference respiratory phase $j_{ref}$ and for the training of the breathing model.

The respiratory signal used for model training is shown in Fig.~\ref{sfig:spiro}, gray curve. We show the areas with plausible breathing simulation and use the unscaled respiratory signal of $pat3D$ with larger variance to provoke artifacts (Fig.~\ref{sfig:spiro}, blue curve). The model training according to Eqs. \ref{eq:model4D} and \ref{eq:model3D} is very efficient using matrix computations.

{\color{black}We use manual expert segmentations of the liver and lungs, available for every phase of the 4D patient, to mainly assess the quality of the inter-patient registration in Eq.~\ref{eq:inter}. Via the availabe inter-phase registrations $\varphi^{pat4D}_j$  (Eq.~\ref{eq:intra}) to the 4D reference phase, we first warp the phase segmentation masks accordingly. After applying the inter-patient registration to $pat3D$, we have the segmentation masks of $pat4D$ in the space of the targeted 3D patient. Now for this patient, also a manual expert segmentation is availabe for comparison.
Quantitatively, the DICE coefficients of the transferred segmentation masks (liver, lungs) can be given to classify the quality of the registration chain of the reference respiratory phases (single atlas approach). 
Qualitatively, we present sample images for four time instants and a movie.

The mean DICE coefficients of the single-atlas registration of the liver and lung masks to the new static patient $ pat3D $ yield satisfying values of 0.86$\pm$0.12 and 0.96$\pm$0.09. Note the clearly different scan ranges of the data sets (Fig.~\ref {G53-img:FoV-motionsig}a).
The animation of the relevant structures is shown as an example in Fig.~\ref {G53-img:motionct4D}, using a variable real breathing signal of the target patient $ pat3D $ (Fig.~\ref {G53-img:FoV-motionsig}b). In the puncture-relevant liver region, the patient's breathing states are simulated plausibly for the 4D reference patient (Fig.~\ref {G53-img:motionct4D}) and, more importantly, the 3D patient (Figs. \ref {G53-img:motionct3D}, \ref {G53-img:motionArt}), to which the motion model of $pat4D$ was transferred\footnote{\href{https://goo.gl/DVVYzw}{Demo movie, click here}}.

\section{DISCUSSION, OUTLOOK AND CONCLUSION}
{\color{black}For interested readers, the basic techiques for 4D breathing motion models have been introduced in \cite{G53-16} by our group. However there, the motion model is restricted to the inside of the lungs and by design a mean motion model is built from several 4D patients. The mean motion model is artificial to some degree, more complex and timely to build.}
The method described here for the transfer of retrospectively modeled respiratory motion of one 4D reference patient to a new 3D patient data set is less complex and extends to a larger body area. It already allows the plausible animation of realistic respiratory movements in a 4D-VR-training-simulator with visuo-haptic interaction. Of course in the future, we want to build a mean motion model for the whole body section including (lower) lungs and the upper abdomen, too.

{\color{black}
In other studies, we found $\alpha_{D}=1$ in Eq.~\ref{eq:inter} robust (compromise between accuracy and smoothness) for inter-patient registration with large shape variations  \cite{G53-10,G53-09}. In Eq.~\ref{eq:intra} for intra-patient inter-phase registration, we use $\alpha_{S}=0.1$ to allow more flexibility for more accuracy as the shape variation between two phases of the same patient is much smaller \cite{G53-14}.}

We achieve qualitatively plausible results for the liver area in this feasibility study. In the upper thorax especially at the rib cage in neighborhood to the dark lungs stronger artifacts can occur (Fig.~\ref{sfig:ribArt}). They are due to problems in the inter-patient registration that is a necessary step for the transfer of the motion model. The non-linear deformation sometimes is prone to misaligned ribs.
The same is true for the lower thorax with perforation first of the liver and then diaphragm (Fig.~\ref {sfig:diaArt}). Further optimization have to be carried out as artifacts can appear on the high contrast lung edge (diaphragm, ribs) with a small tidal volume.
For liver punctures only, the artifacts of smeared ribs are minor as can be seen in Fig.~\ref {G53-img:motionct3D}.

Summing up, the previous assumption from Sec.~\ref{sec:recent} of a dose-relevant or expensive acquisition of a 4D-CT data set for each patient, can be mitigated for liver punctures by the presented transfer of an existing 4D breathing model.

Future work will deal with the better adaptation and simulation of the breathing signal.
Further topics are the optimization of the inter-patient registration and the construction of alternatively selectable mean
4D reference breathing models. {\color{black}As in \cite{G53-01}, the authors plan to perform usability studies with medical practitioners.}

To conclude, the method allows VR needle puncture training in the hepatic area of breathing virtual patients based on a low-risk and cheap 3D data acquisition for the new patient only.
The requirement of a dose-relevant or expensive acquisition of a 4D CT data set for each new patient can be mitigated by the presented concept.
Future work will include the reduction of artifacts and building mean reference motion models.

\section{ACKNOWLEDGEMENT}
Support by grant: DFG HA 2355/11-2.
\bibliographystyle{wscg-alpha}
\if\anon0
\bibliography{G53}
\else
\bibliography{G53-NN}

\begin{thebibliography}{99}

\bibitem[Ehr11]{G53-16}
Ehrhardt, J., Werner, R., Schmidt-Richberg, A., Handels, H.
\newblock Statistical modeling of {4D} respiratory lung motion using
  diffeomorphic image registration.
\newblock {IEEE Transactions on Medical Imaging}, 30(2):251--265, September
  2011.

\bibitem[For12]{G53-06}
Fortmeier, D., Mastmeyer, A., Handels, H.
\newblock {GPU}-based visualization of deformable volumetric soft-tissue for
  real-time simulation of haptic needle insertion.
\newblock {German Conference on Medical Image Processing BVM - 2012: Algorithms
  - Systems - Applications. Proceedings from 18.-20. March 2012 in Berlin},
  pages 117--122, 2012.

\bibitem[For13]{G53-05}
Fortmeier, D., Mastmeyer, A., Handels, H.
\newblock Image-based palpation simulation with soft tissue deformations using
  chainmail on the {GPU}.
\newblock {German Conference on Medical Image Processing - BVM 2013}, pages
  140--145, 2013.

\bibitem[For14]{G53-12}
Fortmeier, D., Mastmeyer, A., Handels, H.
\newblock An image-based multiproxy palpation algorithm for patient-specific
  {VR}-simulation.
\newblock {Medicine Meets Virtual Reality 21, MMVR 2014}, pages 107--113, 2014.

\bibitem[For15]{G53-01a}
Fortmeier, D., Wilms, M., Mastmeyer, A., Handels, H.
\newblock Direct visuo-haptic {4D} volume rendering using respiratory motion
  models.
\newblock {IEEE Trans Haptics}, 8(4):371--383, 2015.

\bibitem[For16]{G53-01}
Fortmeier, D., Mastmeyer, A., Schr{\"o}der, J., Handels, H.
\newblock A virtual reality system for {PTCD} simulation using direct
  visuo-haptic rendering of partially segmented image data.
\newblock {IEEE J Biomed Health Inform}, 20(1):355--366, 2016.

\bibitem[Mas13]{G53-10}
Mastmeyer, A., Fortmeier, D., Maghsoudi, E., Simon, M., Handels, H.
\newblock Patch-based label fusion using local confidence-measures and weak
  segmentations.
\newblock {Proc. SPIE Medical Imaging: Image Processing}, pages 86691N--1--11,
  2013.

\bibitem[Mas14]{G53-04}
Mastmeyer, A., Hecht, T., Fortmeier, D., Handels, H.
\newblock Ray-casting based evaluation framework for haptic force-feedback
  during percutaneous transhepatic catheter drainage punctures.
\newblock {Int J Comput Assist Radiol Surg}, 9:421--431, 2014.

\bibitem[Mas16]{G53-09}
Mastmeyer, A., Fortmeier, D., Handels, H.
\newblock Efficient patient modeling for visuo-haptic {VR} simulation using a
  generic patient atlas.
\newblock {Comput Methods Programs Biomed}, 132:161--175, 2016.

\bibitem[Mas17]{G53-20}
Mastmeyer, A., Fortmeier, D., Handels, H.
\newblock Evaluation of direct haptic 4d volume rendering of partially
  segmented data for liver puncture simulation.
\newblock {Nature Scientific Reports}, 7(1):671, 2017.

\bibitem[Nic05]{G53-18}
Nicolau, S., Pennec, X., Soler, L., Ayache, N.
\newblock A complete augmented reality guidance system for liver punctures:
  First clinical evaluation.
\newblock {Medical Image Computing and Computer-Assisted Intervention--MICCAI
  2005}, pages 539--547, 2005.

\bibitem[Rei06]{G53-19}
Reitinger, B., Bornik, A., Beichel, R., Schmalstieg, D.
\newblock Liver surgery planning using virtual reality.
\newblock {IEEE Computer Graphics and Applications}, 26(6):36--47, 2006.

\bibitem[Sep02]{G53-07}
Seppenwoolde, Y., Shirato, H., Kitamura, K., Shimizu, S., Herk, M.van ,
  Lebesque, J.~V., Miyasaka, K.
\newblock Precise and real-time measurement of {3D} tumor motion in lung due to
  breathing and heartbeat, measured during radiotherapy.
\newblock {Int J Radiation Oncololgy, Biology, Physics}, 53(4):822--834, Jul
  2002.

\bibitem[SR12]{G53-14}
Schmidt-Richberg, A., Werner, R., Handels, H., Ehrhardt, J.
\newblock Estimation of slipping organ motion by registration with
  direction-dependent regularization.
\newblock {Medical Image Analysis}, 16(1):150 -- 159, 2012.

\bibitem[Thi98]{G53-17}
Thirion, J.-P.
\newblock Image matching as a diffusion process: an analogy with maxwell's
  demons.
\newblock {Medical Image Analysis}, 2(3):243 -- 260, 1998.

\bibitem[Vil11]{G53-03}
Villard, P., Boshier, P., Bello, F., Gould, D.
\newblock Virtual reality simulation of liver biopsy with a respiratory
  component.
\newblock {Liver Biopsy, InTech}, pages 315--334, 2011.

\bibitem[Vil14]{G53-02}
Villard, P., Vidal, F., Cenydd, L., Holbrey, R., Pisharody, S., Johnson, S.,
  Bulpitt, A., John, N., Bello, F., Gould, D.
\newblock Interventional radiology virtual simulator for liver biopsy.
\newblock {Int J Comput Assist Radiol Surg}, 9(2):255--267, 2014.

\bibitem[Wil14]{G53-11}
Wilms, M., Werner, R., Ehrhardt, J., et~al.
\newblock Multivariate regression approaches for surrogate-based diffeomorphic
  estimation of respiratory motion in radiation therapy.
\newblock {Phys Med Biol}, 59:1147{\textendash}1164, 2014.

\end{thebibliography}
\fi 

\end{document}